\pdfoutput=1

\documentclass[11pt]{article}
\usepackage[]{EMNLP2023}

\usepackage{times}
\usepackage{float}
\usepackage{latexsym}
\usepackage{amsmath}
\usepackage[T1]{fontenc}
\usepackage{amsmath} 
\DeclareUnicodeCharacter{03BB}{\ensuremath{\lambda}}

\usepackage[utf8]{inputenc}
\usepackage{graphicx} 
\usepackage{epstopdf}
\usepackage{microtype}

\usepackage{inconsolata}

\title{On the token distance modeling ability of higher RoPE attention dimension}

\author{
 \textbf{Xiangyu Hong$^{1}$\Thanks{~Equal contribution.}},
 \textbf{Che Jiang$^{1*}$\Thanks{~The work was done when Che Jiang worked as intern at Pattern Recognition Center, WeChat AI, Tencent Inc, China.}}, \textbf{Biqing Qi$^{1}$\Thanks{~Corresponding authors}}\\
 \textbf{Fandong Meng$^{2}$}, \textbf{Mo Yu$^{2}$}, \textbf{Bowen Zhou$^{1\ddag}$}, \textbf{Jie Zhou$^{2}$}
\\
$^1$ Department of Electronic Engineering, Tsinghua University\\
$^2$ Pattern Recognition Center, WeChat AI, Tencent Inc, China \quad 
\\
\texttt{hong-xy22@mails.tsinghua.edu.cn} \texttt{jc23@mails.tsinghua.edu.cn} \\ \texttt{zhoubowen@tsinghua.edu.cn}
\\
\\
}

\begin{document}
\maketitle
\begin{abstract}
Length extrapolation algorithms based on Rotary position embedding (RoPE) have shown promising results in extending the context length of language models. However, understanding how position embedding can capture longer-range contextual information remains elusive. 
Based on the intuition that different dimensions correspond to different frequencies of changes in RoPE encoding, we conducted a dimension-level analysis to investigate the correlation between a hidden dimension of an attention head and its contribution to capturing long-distance dependencies.
Using our correlation metric, we identified a particular type of attention heads, which we named \emph{Positional Heads}, from various length-extrapolated models.
These heads exhibit a strong focus on long-range information interaction and play a pivotal role in long-input processing, as evidenced by our ablation.
We further demonstrate the correlation between the efficiency of length extrapolation and the extension of the high-dimensional attention allocation of these heads.
The identification of Positional Heads provides insights for future research in long-text comprehension.

\end{abstract}

\section{Introduction}

The Transformer model has revolutionized natural language processing tasks, but it demonstrates limitations in modeling long sequences. Meanwhile, models like Mamba \cite{gu2023mamba} that excel in capturing long-range dependencies struggle to meet the practical requirements of natural language modeling \cite{lieber2024jamba}. Consequently, there has been a recent surge of work focused on extending the context length in language models based on the Transformer architecture \cite{zhang2024soaring,xiong2023effective,Fu2024DataEF}. Particularly, some of these efforts that leverage and enhance the capabilities of RoPE (Rotary Positional Embedding) \cite{Jin2024LLMML,Peng2023YaRNEC,Chen2023CLEXCL}, have shown promising results in extrapolating the model's capacity to handle longer contexts \cite{wang2024beyond}.

Open-source large language models commonly employ Rotary Positional Embedding (RoPE) to model sequence positional information \cite{Touvron2023LlamaOA,jiang2023mistral,yang2023baichuan,bai2023qwen}. RoPE exhibits two desirable properties. Firstly, its exponential positional encoding introduces long-range attention decay, allowing the model to focus more on neighboring semantic information. Secondly, by utilizing trigonometric functions to differentiate frequencies, RoPE effectively captures different distances between tokens, enabling higher attention scores for tokens with longer semantic dependencies, facilitating semantic aggregation. When compared to length extrapolation methods based on sparse attention \cite{ratner2022parallel, Xiao2023EfficientSL} or prompt compression \cite{yen2024long, xiao2024infllm}, modifications to RoPE for length extrapolation do not result in the loss of fine-grained contextual information at a global level. Therefore, it possesses distinct advantages in tasks such as long text comprehension \cite{Bai2023LongBenchAB,lv2024longwanjuan}, where the preservation of comprehensive contextual information is essential for practical applications.

A prevailing viewpoint suggests that language models based on RoPE encounter out-of-distribution (OOD) issues when faced with contexts longer than the pre-training text length, specifically affecting the sampling of the trigonometric function component for token distances \cite{Peng2023YaRNEC,xiong2023effective}. As a result, related studies have adjusted the attention resolution in the context of long texts and fine-tuned the model to adapt to longer token distances. We hypothesize that the effectiveness of such methods stems from RoPE's ability to decouple information from different distances by representing them through different dimensions with varying rotational frequencies. However, this pattern has not been thoroughly observed and analyzed in the inference process.

Our paper presents a novel approach by examining the impact of each dimension of attention heads that use RoPE on modeling text distances. We empirically validate the claim in Yarn that lower-frequency dimensions are responsible for modeling longer text dependencies. Furthermore, we discover that not all attention heads exhibit this characteristic, emphasizing the importance of heads that possess this relationship in modeling long texts. Our study explores RoPE's potential for long text modeling from a frequency perspective, shedding light on the relationship between dimensions and text modeling capabilities. Our primary findings are as follows:
\begin{itemize}
    \item In most attention heads, regardless of whether length extrapolation is performed, the impact of high-dimensional low-frequency components is greater than that of low-dimensional high-frequency components.
    \item Input lengths exceeding the pre-training length can result in anomalies in high-dimensional components. Length extrapolation extends the high-dimensional attention allocation for a longer token distance.
    \item We refer to attention heads with stronger correlation between token distance and dimension allocation as \emph{Positional Heads}, which play a crucial role in modeling text distances.
\end{itemize}


\section{Background}

\subsection{Rotary Position Embeddings}

Large Language Models (LLMs) are primarily based on the Transformer architecture \cite{Vaswani2017AttentionIA}, with the attention mechanism at its core. A prevalent method for incorporating positional information in these models is Rotary Position Embeddings (RoPE) \cite{Su2021RoFormerET}, which leverages rotation matrices to encode the positional information of sequences.

In RoPE, the positional encoding for a hidden layer, with the hidden dimension denoted by \(d\), uses a rotation matrix for each position \(m\). The rotation matrix \(\mathcal{R}_m\) is defined as follows:


\begin{equation}
\text{\includegraphics[width=\linewidth]{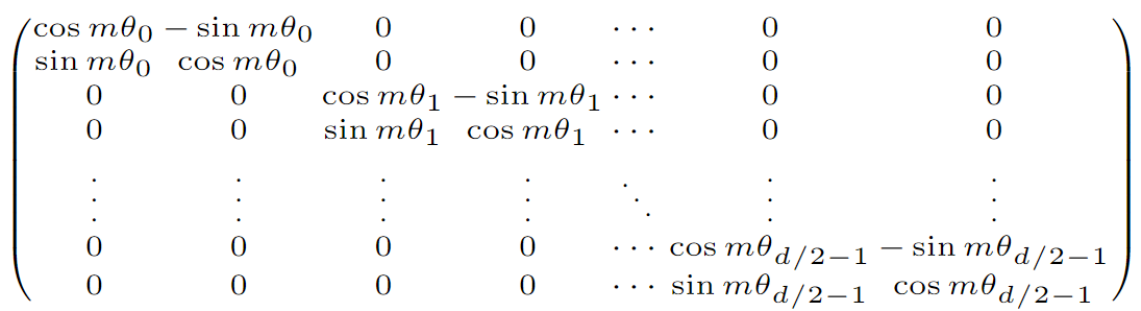}}
\end{equation}

where

\begin{equation}
\theta_i = 10000^{-2i/d}
\end{equation}

Explicitly, for the query vector \(\mathbf{q}\) at position \(m\) and the key vector \(\mathbf{k}\) at position \(n\), we have:

\begin{equation}
\mathbf{q} = \begin{bmatrix} q_{0} \\ q_{1} \\ \vdots \\ q_{d-1} \end{bmatrix}, 
\quad \mathbf{k} = \begin{bmatrix} k_{0} \\ k_{1} \\ \vdots \\ k_{d-1} \end{bmatrix}
\end{equation}

After applying RoPE, the transformed vectors \(\mathbf{q}_m\) and \(\mathbf{k}_n\) are given by:
\begin{equation}
\mathbf{q}_m = \mathcal{R}_m \mathbf{q} =
\begin{bmatrix} q_{m0} \\ q_{m1} \\ \vdots \\ q_{m{(d-1)}} \end{bmatrix},
\end{equation}
\begin{equation}
\mathbf{k}_n = \mathcal{R}_n \mathbf{k} =
\begin{bmatrix} k_{n0} \\ k_{n1} \\ \vdots \\ k_{n{(d-1)}} \end{bmatrix}    
\end{equation}

The attention weights are then calculated using the dot product of the transformed vectors and the dot product for $\mathbf{q}_m$ and $\mathbf{k}_n$ is given by:
\begin{equation}
\mathbf{q}_m^T \mathbf{k}_n = \sum_{i=0}^{d-1} q_{m,i} k_{n,i}    
\end{equation}

\subsection{Length Extrapolation Methods}

We have investigated methods to extend the context length of language models, particularly using Rotary Position Embedding (RoPE). Our research focuses on three prominent techniques: Yarn \cite{Peng2023YaRNEC}, CLEX \cite{Chen2023CLEXCL}, and SelfExtend \cite{Jin2024LLMML}. Each method leverages different aspects of positional encoding to enhance long-range token interactions, showing favorable performance in our tests.

\textbf{YaRN} \cite{Peng2023YaRNEC} addresses Out-of-Distribution (OOD) scenarios by categorizing RoPE dimensions into three frequency-based groups and applying tailored interpolation strategies. Low-frequency dimensions use linear interpolation with adjusted $\theta_i$ (λ=1) for smooth transitions. High-frequency dimensions remain unchanged, while intermediate-frequency dimensions use linear interpolation to bridge the extremes effectively.

\textbf{CLEX} \cite{Chen2023CLEXCL} advances the concept of Dynamic Scaling by modeling $\theta_i(\text{pos})$ as a continuous function of position using a neural ODE. This method enables precise parameter fine-tuning to fit the data, demonstrating superior performance in our tests.

\textbf{SelfExtend} \cite{Jin2024LLMML} uses bi-level attention: grouped attention and neighbor attention, to capture dependencies among both distant and adjacent tokens. It addresses positional O.O.D. issues by remapping unseen large relative positions to those encountered during pretraining through a floor division operation. This approach allows LLMs to maintain coherence over longer texts without fine-tuning.

\section{Defining Dimension Contribution in RoPE}

In Rotary Position Embedding (RoPE), each dimension of the vectors \(\mathbf{q}_m\) and \(\mathbf{k}_n\) contributes to the attention score via their dot product. To thoroughly investigate the role of different dimensions in RoPE for semantic modeling, we utilize an algorithm that analyzes the contribution of each dimension to the attention scores.

To capture the contribution of each dimension, we employ the Hadamard product, i.e., element-wise multiplication, denoted by the symbol $\odot$:

\begin{equation}
\mathbf{h} = \mathbf{q}_m \odot \mathbf{k}_n \in \mathbf{R}^d , \mathbf{h}_i = q_{m,i} k_{n,i},
\end{equation}
where
\begin{align}
 {h}_{2i} = & \,  {q}_{2i}  {k}_{2i} \cos(m \theta_i) \cos(n \theta_i) \nonumber \\
 & \, -  {q}_{2i+1}  {k}_{2i} \sin(m \theta_i) \cos(n \theta_i) \nonumber \\
 & \, -  {q}_{2i}  {k}_{2i+1} \cos(m \theta_i) \sin(n \theta_i) \nonumber \\
 & \, +  {q}_{2i+1}  {k}_{2i+1} \sin(m \theta_i) \sin(n \theta_i) \nonumber\\
\end{align}
\begin{align}
{h}_{2i+1} = &\ {q}_{2i} {k}_{2i} \sin(m \theta_i) \sin(n \theta_i) \nonumber \\
        & \, + {q}_{2i+1}  {k}_{2i} \cos(m \theta_i) \sin(n \theta_i) \nonumber \\
        & \, +  {q}_{2i}  {k}_{2i+1} \sin(m \theta_i) \cos(n \theta_i) \nonumber \\
        & \, +  {q}_{2i+1}  {k}_{2i+1} \cos(m \theta_i) \cos(n \theta_i)
\end{align}

In RoPE, every two dimensions correspond to trigonometric functions with the same frequency $\theta_i$. We sum the values of these corresponding dimensions to form new vectors:
\begin{equation}
\mathbf{g} \in \mathbf{R}^\frac{d}{2}, \quad g_i = h_{2i} + h_{2i+1}  
\label{g_compute}
\end{equation}
for \( i = 0, 1, 2, \ldots, \frac{d}{2} - 1 \).

The value of \( g_{i} \) reflects the contribution of each dimension in RoPE to the attention score. A higher value indicates a greater contribution of that dimension to the attention score, where:
\begin{equation}
\begin{split}
g_{i} &= h_{2i} + h_{2i+1} \\
&= (q_{2i} k_{2i} + q_{2i+1} k_{2i+1}) \cos((m-n) \theta_i) \\
&\quad + (q_{2i} k_{2i+1} - q_{2i+1} k_{2i}) \sin((m-n) \theta_i).
\end{split}
\end{equation}
Here, \( \theta_i \) represents the positional encoding frequency for the \( i \)-th dimension.

The dot product of \( \mathbf{q}_m \) and \( \mathbf{k}_n \) can be expressed as:
\begin{equation}
\mathbf{q}_m^T \mathbf{k}_n = \sum_{i=0}^{d-1} q_{m,i} k_{n,i} = \sum_{i=0}^{d-1} h_{i} = \sum_{i=0}^{\frac{d}{2}-1} g_{i}
\end{equation}
Therefore, we use the value \( g_i \) to measure the contribution of \( \theta_i \) to the attention score.

This methodological framework enables a comprehensive analysis of how each dimension in Rotary Position Embedding (RoPE) contributes to the attention scores. Through this approach, we can delve into the role of different dimensions in RoPE for semantic modeling.

\section{Experiments}
\subsection{Study on dimension-level contributions to attention scores}
This study aims to answer the following question:\\
\emph{\textbf{Are there distinct patterns of attention contributions across different dimensions?}}

\begin{figure}[h] 
    \includegraphics[width=\columnwidth]{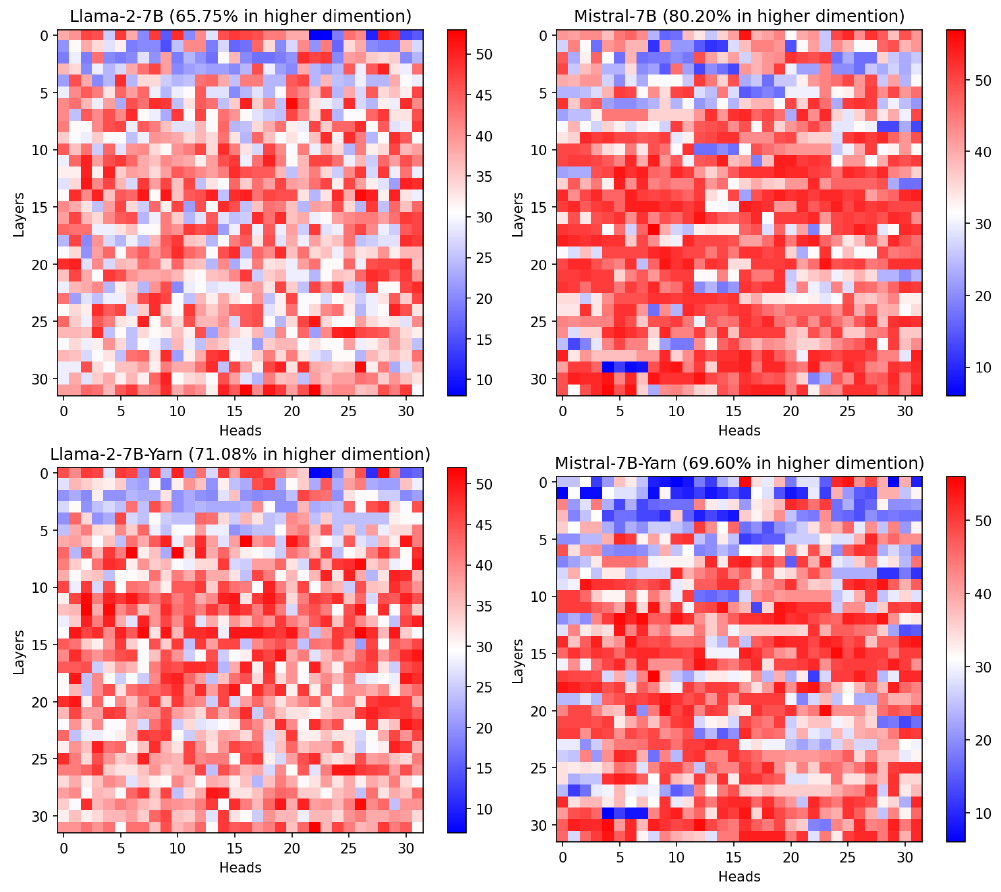} 
    \caption{The average of the dimensional distribution of attention scores for each head in each layer of the four models}
    \label{fig:dim_heatmap}
\end{figure}

To examine this, we initially observed the overall contribution of each dimension to the attention scores. We sampled 17 inputs, and for each input, at each layer and each head of the model, we randomly selected $100 \times \text{number of tokens}$ $qk$ pairs. For each selected $qk$ pair, we computed the contribution of each dim as shown in Section 3, then recorded the top 5 dimensions that contributed the most. We conducted a statistical analysis of the distribution of attention scores in terms of dimensions for each layer and head across four models: the original Llama-2-7B, Mistral-7B, and their versions with 64K length extrapolation using the Yarn method. The average values of these dimension distributions are presented in Figure \ref{fig:dim_heatmap}. As per the theoretical analysis of RoPE, the models tend to focus on syntactic parsing in the shallow layers, placing greater emphasis on shorter-distance information. The attention scores of the majority of attention heads are predominantly contributed by the higher-dimensional components. There were no significant changes observed in the dimension distribution before and after length extrapolation.

\begin{figure}[h] 
    \includegraphics[width=\columnwidth]{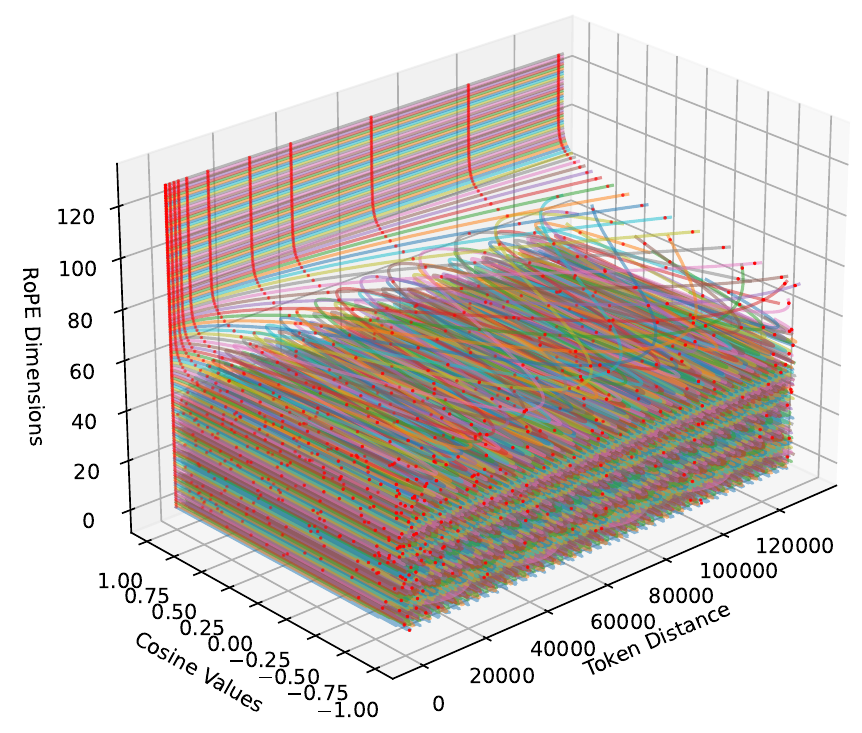} 
    \caption{The value of trigonometric function in rotation position coding changes with the dimension and token distance, and the red dots represents the trigonometric function value of different dimension corresponding to a specific token distance}
    \label{fig:high_explain}
\end{figure}

We now provide a potential explanation for the higher contribution observed in the higher-dimensional components. According to Equation (11), the effect of the rotated positional encoding matrix on positions n and m in dimension i is equivalent to a trigonometric function of the form $cos(n-m)\theta_i$. We visualized the values of these trigonometric functions, as shown in Figure \ref{fig:high_explain}. The red dots represent several distances (n-m) with the maximum token distance set at 128k. It can be observed that tokens with longer distances correspond to shorter distinguishable curves in the rotated positional encoding. In the lower-dimensional range, the abrupt changes in values between adjacent dimensions become irregular due to the higher frequency of the trigonometric function. The purpose of RoPE is to encode varying token distances across different dimensions. The aggregation of information from these different dimensions is carried out when computing attention scores.The irregularity in the lower-dimensional range hinders the disentanglement of distance-related information. Consequently, during the training process, the model tends to favor the working of attention in the higher-dimensional components. Moreover, the maximum text length that the model can handle is also determined by the higher-dimensional components. Furthermore, increasing the base of the exponential function lowers the frequency of the trigonometric function, leading to increased distinguishable components in the higher dimensions. This has been confirmed to be a practical method for length extrapolation in pre-training approaches such as Llama3\cite{llama3modelcard} and Code-Llama \cite{roziere2023code}.
\begin{figure*}[t] 
    \includegraphics[width=2\columnwidth]{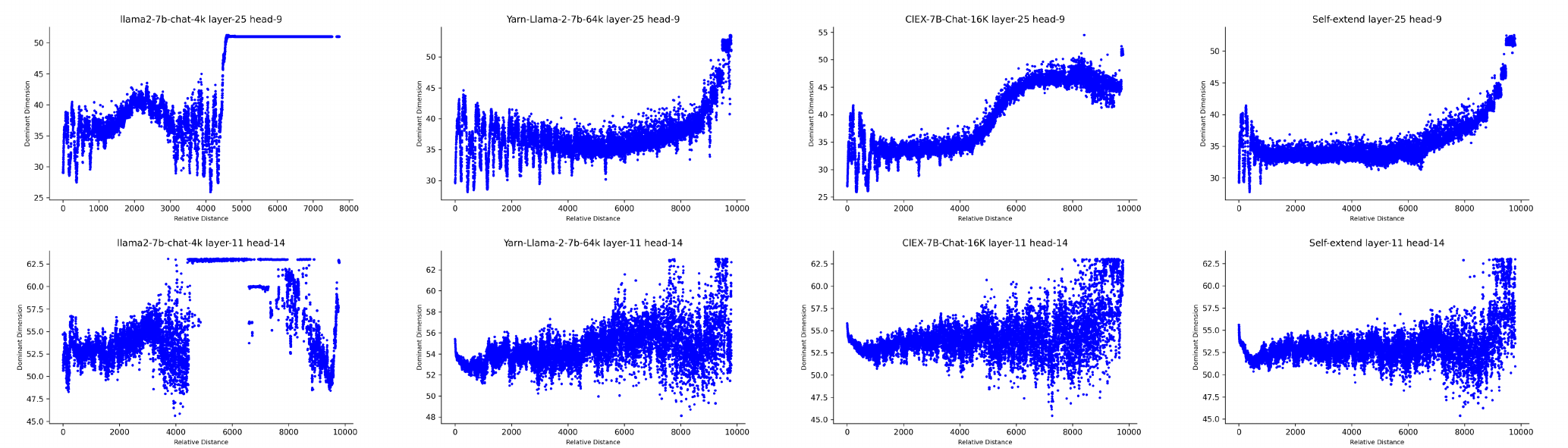} 
    \caption{Correlation plot comparing the original Llama model with three different length extrapolation methods. The top and bottom rows show correlations in different heads.}
    \label{fig:dimension}
\end{figure*}
\subsection{Study on correlation between dimensions and token distances}
The previous study confirms the significant contribution from higher dimensions. Continued from the conclusion, in this study, we aim to understand:
\emph{\textbf{Are higher dimensions responsible for long-range attention among tokens?}}

\subsubsection{Correlation Plot}
Indeed, according to the principles of RoPE, higher dimensions are responsible for modeling longer token distances. However, it remains to be examined whether this correlation strictly holds in the actual inference process of pre-trained models such as Llama. To investigate this, we primarily focused on Llama and employed the methods shown below to observe the original Llama model as well as three different length extrapolation methods.
To comprehensively assess the influence of all dimensions in Rotary Position Embedding (RoPE) on a given query-key pair ($\mathbf{q_m}$ and $\mathbf{k_n}$), we propose an algorithm to compute the \textit{Dominant Dimension}. This value is determined by analyzing the contribution scores assigned to each dimension within RoPE. The dominant dimension signifies that the attention score predominantly originates from the vicinity of this particular dimension.
For each vector $\mathbf{g}_i$ in \eqref{g_compute}, we apply the softmax function:
\begin{equation}
\text{softmax}(\mathbf{g})_i = \frac{e^{\mathbf{g}_i}}{\sum_{j} e^{\mathbf{g}_j}}    
\end{equation}
We then compute the dot product of the softmax output with its corresponding position vector to determine the dominant dimension:
\begin{equation}
\text{Dominant Dimension} = \text{softmax}(\mathbf{g}) \cdot \mathbf{pos}    
\end{equation}
where 
\begin{equation}
\mathbf{pos}=  \begin{bmatrix} 0 & 1 & \ldots & \frac{d}{2}-1 \end{bmatrix}^T
\end{equation}

To investigate the relationship between relative distance and dominant dimension, we sampled 17 prompts. For each prompt, across every layer and head of the model, we selected the top 100 tokens with the highest interaction attention scores for each token. This resulted in 100 times the number of tokens qk pairs. For each qk pair, dominant dimension was computed, and its relative distance \(m-n\) was recorded.

For each head, we obtained a collection of \(100 \times \) number of tokens Relative Distance - Dominant Dimension pairs. If a relative distance corresponds to multiple dominant dimensions, we averaged them to obtain the dominant dimension corresponding to that distance. 

The correlation between token relative distances and the dominant dimension of attention scores is depicted in Figure \ref{fig:dimension}. 
To ensure the generalizability of our findings, we conducted correlation analyses across multiple datasets, including several Chinese datasets. The results from these analyses are presented in detail in the appendix.

\subsubsection{Observation}
\label{sec:observation}
Through a thorough analysis of the relationship between the dominant dimension and the relative distance of each head of each layer of the model, we have drawn the following inspiring observation:

\begin{enumerate}
    \item In some heads of the model, there is a significant correlation between the dominant dimension and the relative distance, whereas, in other heads, this correlation is not observed.
    \item For the original Llama model, a sudden change in the dominant dimension occurs when the sequence length exceeds the pre-training length (4K). We observed a similar phenomenon in other models, such as Baichuan, as illustrated in Appendix \ref{baichuan_ood}.
    \item For the length extrapolation method, by observing the dominant dimension of the model, it can be seen that this method extends the trend of the dominant dimension within the pre-training length range of Llama to a new length range, thereby achieving length extrapolation. This observation is consistent with the design methodology of these length extrapolation approaches.
\end{enumerate}

\subsubsection{OOD Explanation}

To further elucidate why the original model exhibits a sudden change in behavior when exceeding the pretraining length, we conducted an ablation study on the dimension matrix $R_{m}$ of the rotary position encoding. The results are depicted in the figure.
\begin{figure}[h] 
    \includegraphics[width=\columnwidth]{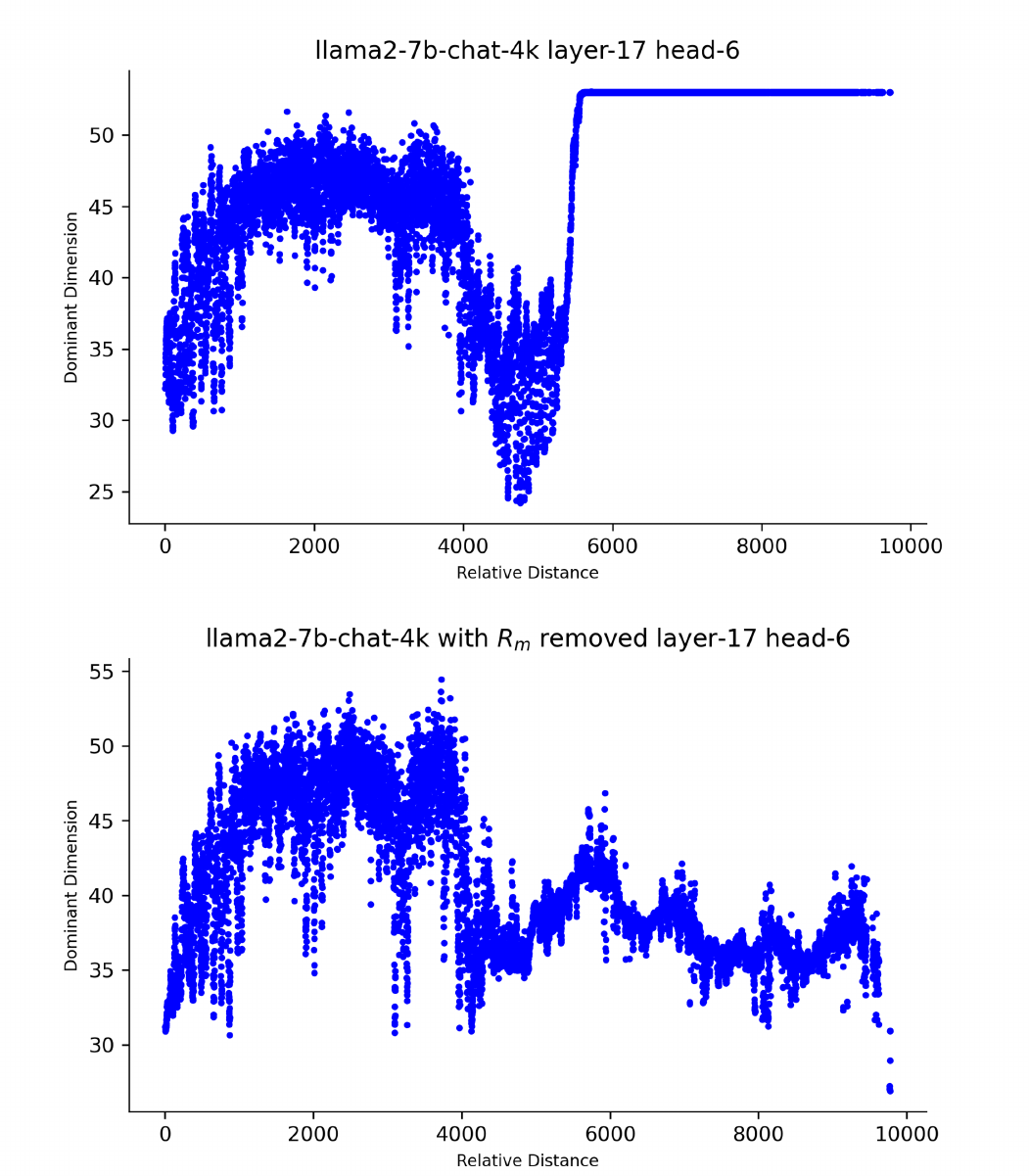} 
    \caption{Correlation plot comparing the original Llama model with the model where \(R_{m}\) has been removed. The above shows the results from the original Llama model, while the below displays the results after removing \(R_{m}\). The abrupt changes disappear after \(R_{m}\) removal.
}
    \label{fig:ood_explain}
\end{figure}
As shown in Figure \ref{fig:ood_explain}, it can be observed that when the length is less than the pretraining length (4K), the image after removing \( R_m \) shows little difference compared to the original. However, beyond the pretraining length, no abrupt changes occur.
Therefore, we propose a plausible explanation based on the finding: As depicted in Figure \ref{fig:high_explain}, as the relative distance increases, the lower dimensions of the rotary positional encoding tend to resemble characteristics similar to random sampling, while the higher dimensions remain comparatively stable. Consequently, when the relative position exceeds the pretraining length (4K), the values in the lower dimensions gradually become overshadowed by noise from the trigonometric functions, whereas the values in the higher dimensions remain intact. The model training adjusts to accommodate this sampling characteristic of trigonometric functions. However, when the relative distance surpasses the model's pretraining length, the model struggles to adapt to this extended sampling range, leading to a scenario where the lower dimensions lose coherence, while the influence of the higher dimensions becomes predominant.

\subsection{Finding the \emph{Positional Heads}}
\subsubsection{Positional Heads Detection}

\emph{Positional Heads} refer to attention heads with significant correlations mentioned in Section \ref{sec:observation}. In order to identify them, we quantified the distance-dimension correlation for each head using the Spearman rank correlation coefficient. This statistical measure was computed based on the visualization provided earlier. A Spearman correlation coefficient closer to 1 (in absolute value) indicates a stronger correlation, with the sign showing the direction. More details are in Appendix \ref{spearman_all}.

\begin{figure}[h]
    \centering
    \includegraphics[width=0.7\columnwidth]{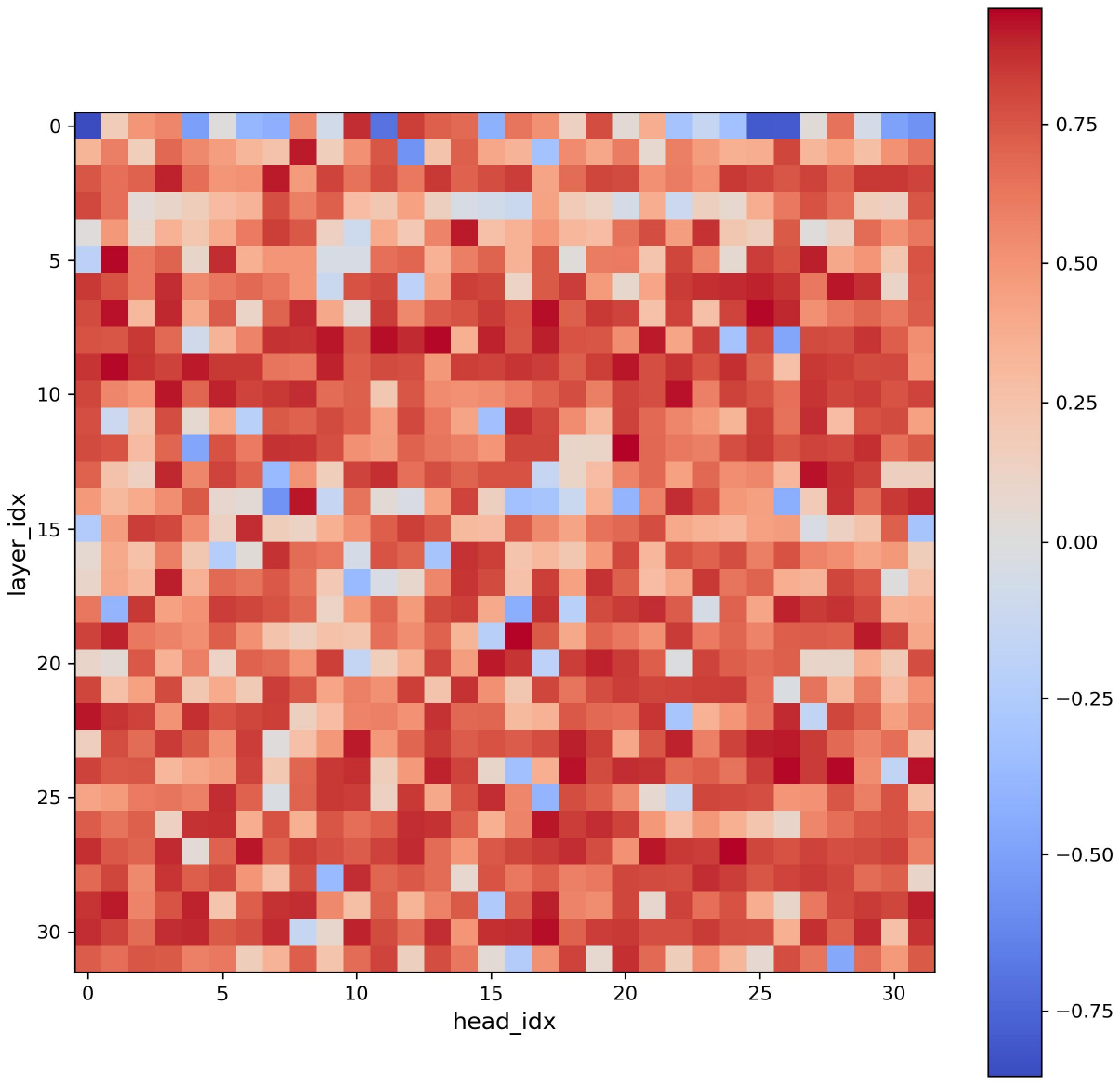}
    \caption{Spearman correlation coefficients of each head in the YaRN-Llama-2-7b-64K model. In most heads, there is a correlation between the dominant dimension and the relative distance.}
    \label{fig:spearman}
\end{figure}

\subsubsection{Influence of Positional Heads on long distance modeling}

To validate the importance of attention heads with high distance-dimension correlations for long text comprehension, we conducted a masking procedure on these heads. Using the metrics described in the previous section, we identified the top 5\% and top 10\% heads based on their rankings and set their output to zero. We then compared the performance of these heads with randomly sampled 5\% and 10\% heads. The results, as shown in the following, demonstrate that heads with high distance-dimension correlations exhibit greater importance across various tasks.

\begin{figure*}[t] 
    \centering
    \includegraphics[width=2\columnwidth]{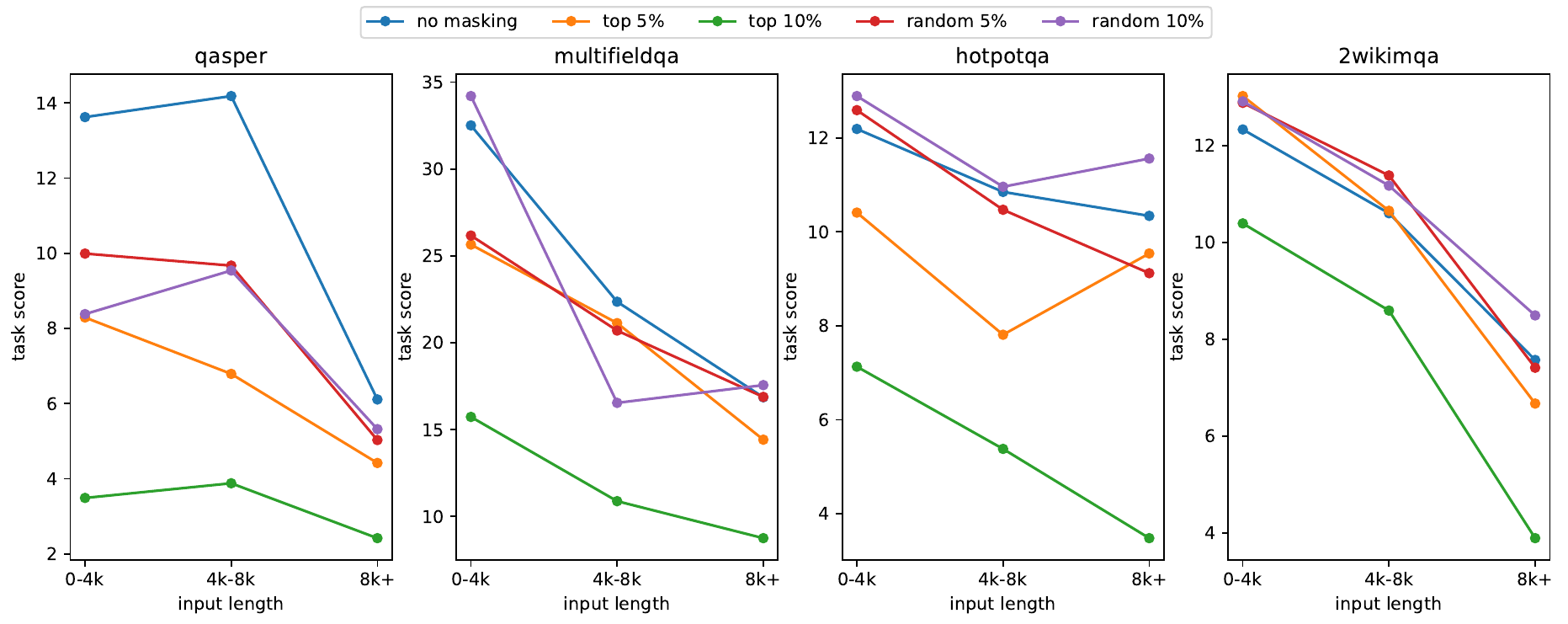} 
    \caption{Masking out top scored heads v.s. random heads. For the QA tasks in LongBench, the removal of heads with top scores clearly reduces performance.}
    \label{fig:yarn_mask}
\end{figure*}

\textbf{Question Answering.} The question-answering (QA) task is a commonly used text comprehension task that requires models to comprehend long text inputs and retrieve information relevant to the given questions. We utilized four QA tasks from LongBench \cite{Bai2023LongBenchAB} to evaluate the impact of random masking of attention heads on the results. Figure \ref{fig:yarn_mask} indicates that randomly masking out attention heads had no significant effect on the results. However, when we masked out the top 5\% and 10\% heads based on the distance-dimension correlation metrics, it resulted in a significant decline in the model's performance on this task. More results can be found in Appendix \ref{other_mask_result}.

\begin{figure}[h] 
    \centering
    \includegraphics[width=\columnwidth]{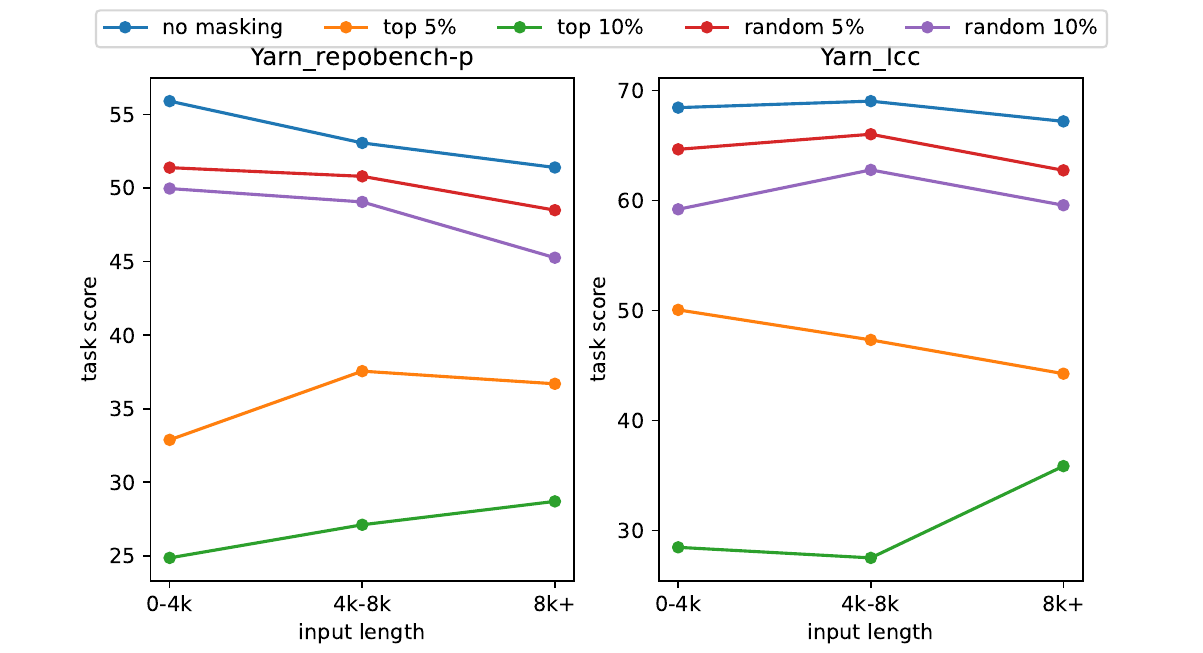} 
    \caption{Masking out top-scored heads v.s. random heads. For the Code tasks in LongBench, the removal of heads with top scores evidently leads to a decline in performance.}
    \label{fig:code_mask}
\end{figure}

\textbf{Code Completion.}
Compared to the QA task, the code completion task places higher demands on long-distance dependencies in the text. We employed the code completion task from LongBench to assess the impact of random masking of attention heads on the results.
As shown in Figure \ref{fig:code_mask}, this observation suggests that our proposed metrics can effectively identify the heads that are more important for understanding long texts from the perspective of long-distance information interaction.

\begin{figure}[h] 
    \includegraphics[width=\columnwidth]{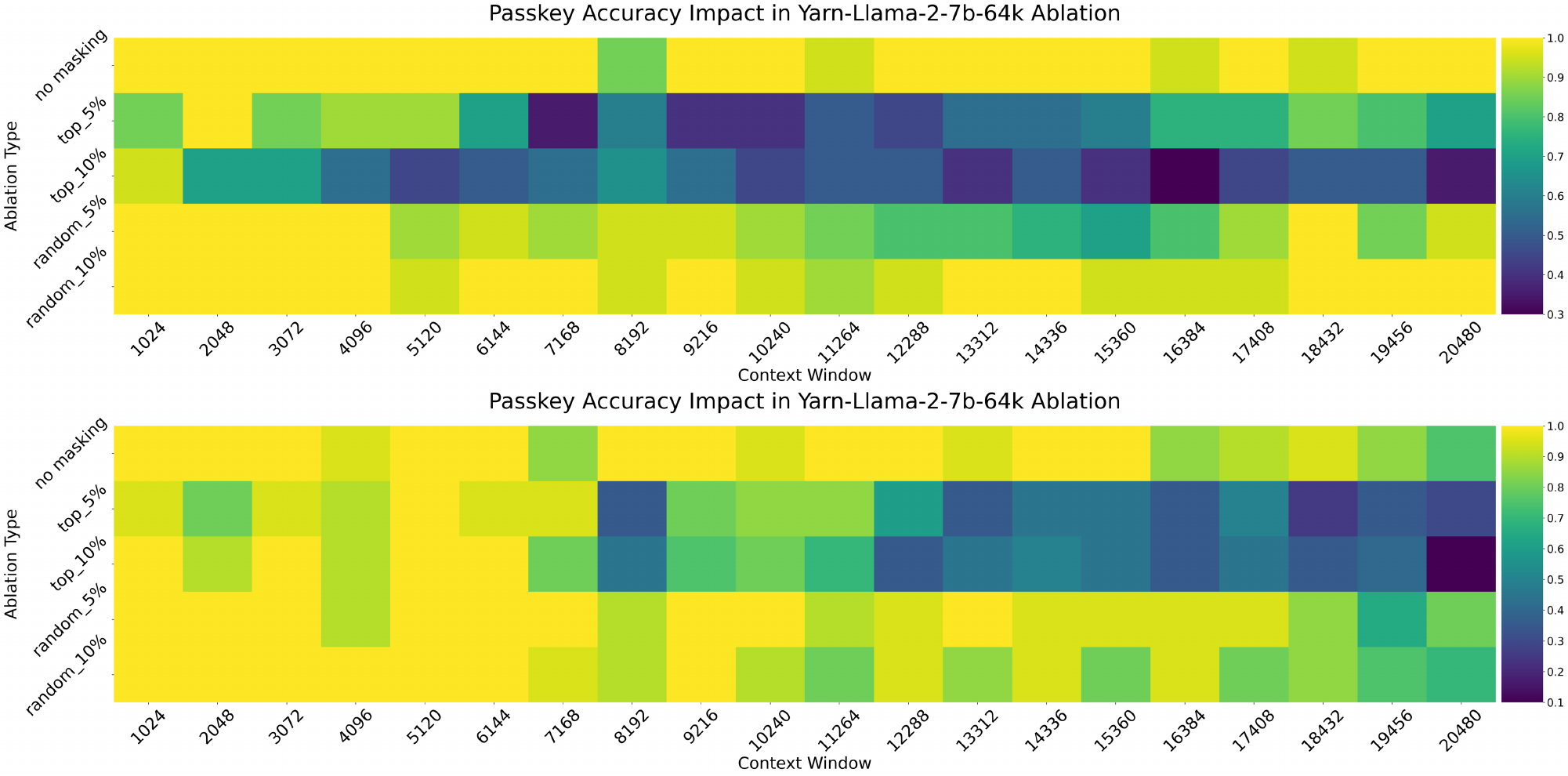} 
    \caption{Masking out heads with top scores v.s. random heads. For the passkey task, the removal of heads with top scores clearly reduces performance.}
    \label{fig:passkey}
\end{figure}

\textbf{PassKey.} The PassKey task is commonly used to evaluate the long text retrieval capability of models. We conducted the same ablation experiments on this task. The models used were the original Llama2-7B model and the Llama2-7B model with length extrapolation using the Yarn method. The results are shown in Figure \ref{fig:passkey}. When the input length exceeds the pre-training length of the model, the original model exhibits out-of-distribution failures in long-distance retrieval. However, when we mask out the high-score attention heads of the length-extrapolated model, the model shows a uniform performance decline across all lengths of retrieval, indicating that these attention heads are highly sensitive to the distance between texts. On the other hand, random masking out of attention heads does not exhibit this phenomenon.

\begin{figure}[h] 
    \includegraphics[width=\columnwidth]{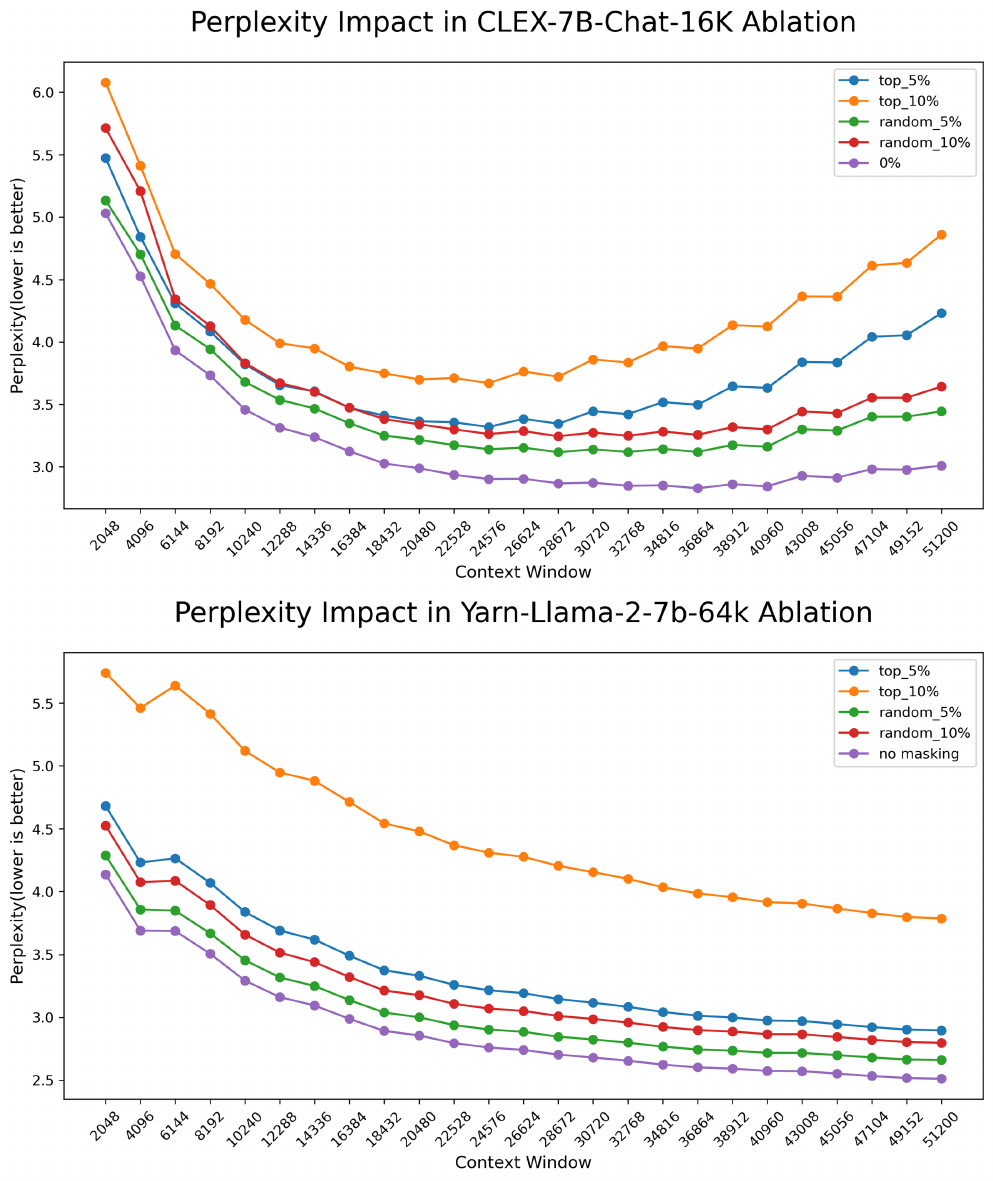} 
    \caption{Masking out heads with top scores v.s. random heads. For the ppl, the removal of heads with top scores clearly reduces performance.}
    \label{fig:ppl}
\end{figure}

\textbf{Perplexity.} While evaluating the long text comprehension ability of the models, it is important to ensure that the fundamental performance of the models does not collapse. We assessed the perplexity (PPL) of the aforementioned models and their ablated versions, and the results are shown in Figure \ref{fig:ppl}. It can be observed that although the ablation of high-score attention heads led to a decrease in PPL, it did not result in the PPL explosion seen in the original Llama model when faced with long texts.

\section{Related Work}
Handling longer contexts in Transformer models has seen significant improvements through various methods, including enhanced training techniques, innovative frameworks, memory mechanisms, and adjustments to positional encoding, including enhanced training techniques\cite{Fu2024DataEF}, external summary designs \cite{Xiao2024InfLLMTL}, memory mechanisms \cite{Dai2019TransformerXLAL,Mohtashami2023LandmarkAR}, and adjustments to positional encoding. Among these methods, modifying positional encoding stands out due to its simplicity and efficacy. Some methods manipulate the token position numbering itself, as seen in PI \cite{Chen2023ExtendingCW} and Selfextend \cite{Jin2024LLMML}. Others make adjustments within the encoding layers at the level of rotational positional encoding, exemplified by works like YaRN \cite{Peng2023YaRNEC} and CLEX \cite{Chen2023CLEXCL}. Additionally, novel positional encodings, such as CoPE \cite{Golovneva2024ContextualPE}, have been proposed to generalize rotational positional encoding and further enhance long-text capabilities.

Certain studies have delved into the impact of positional encoding in depth. Some research indicates that the initial token's position is crucial in long-text contexts \cite{Han2023LMInfiniteSO,Xiao2023EfficientSL}, while other work \cite{Men2024BaseOR} highlights that the base of rotational positional encoding can limit a model's capacity to handle long texts. \citet{Fang2024UniMemTA} proposes a comprehensive framework to describe length extrapolation. However, the role of different dimensions within rotational positional encoding for information interaction remains underexplored. Moreover, the precise mechanisms by which positional encoding affects information interaction are not yet fully understood.

Highly relevant to this work are studies focusing on the interpretability of attention heads\cite{Wu2024RetrievalHM, Olsson2022IncontextLA}. These studies specifically investigate the role of attention heads in information retrieval processes. The function of self-attention mechanisms extends beyond mere replication of highly relevant information; we emphasize the capability of self-attention mechanisms to integrate information from different positions. This capability is crucial for practical long-text comprehension tasks.

\section{Conclusion}
We investigated the properties of attention heads with rotary position embeddings (RoPE) in commonly used Transformer architectures. Using long text comprehension tasks as a starting point, we explored the modeling of token-to-token distance within the model by deconstructing the contributions of different dimensions within the attention heads to the attention scores.

We found that due to the computational nature of rotary position embeddings, higher dimensions of the attention heads, which correspond to lower rotational frequencies, are more effective at distinguishing distances between tokens. Furthermore, attention heads that, through training, allocate attention scores across different dimensions according to token distances and exhibit a certain degree of correlation, demonstrate superior capabilities in modeling text distances. These heads are crucial for integrating information from varying distances in long text comprehension tasks.

We provide an analytical perspective on the currently popular rotary position embeddings, illustrating the attention patterns of models trained with RoPE. Future research can leverage the properties of these attention heads to address challenging tasks such as long text comprehension.

\section*{Limitations}
Although we demonstrated the capability of RoPE in modeling textual distances, several limitations are worth noting. First, our dimensional decomposition approach is based on the explicit meaning of dimensions in rotary position embeddings; this method is not applicable to all types of position encodings. Nonetheless, we maintain that decoupling token distance in attention computation is crucial for integrating and understanding information across different distances. Second, due to computational resource constraints, we could not implement many hypotheses we wished to validate on a larger scale. Our observations were not validated with longer input sequences, and the impact of fine-tuning on these attention heads was not analyzed. We leave a more detailed experimental analysis to future work.

\section*{Acknowledgements}
This work is supported by the National Science and Technology Major Project (No. 2023ZD0121403).
We extend our gratitude to the anonymous reviewers for their insightful feedback.


\bibliography{anthology,custom}
\bibliographystyle{acl_natbib}

\appendix

\section{Appendix}

\label{sec:appendix}
\subsection{OOD phenomenon in other model}
\label{baichuan_ood}
\begin{figure}[h] 
    \includegraphics[width=\columnwidth]{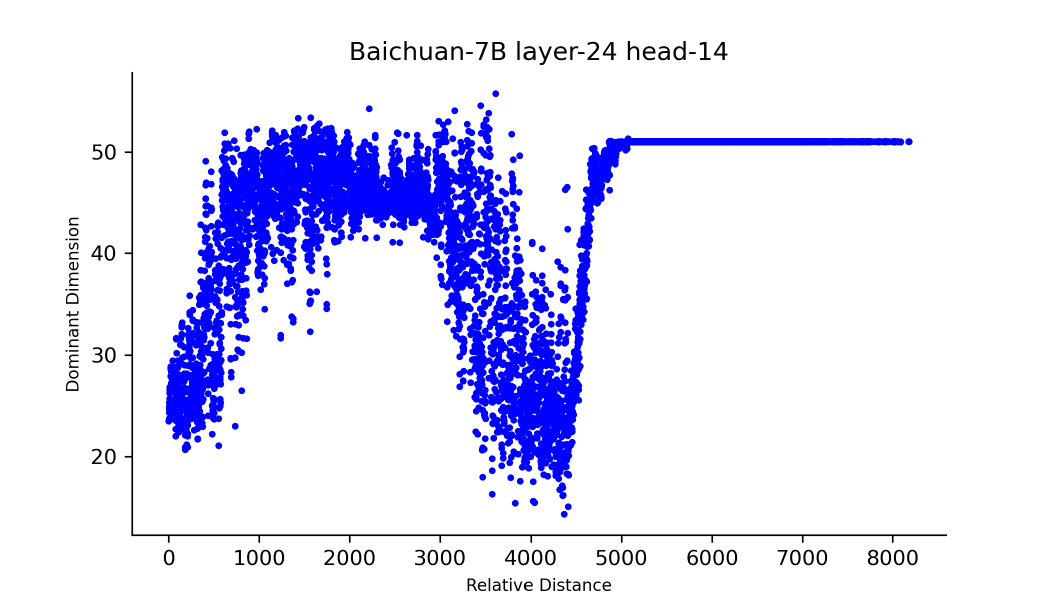} 
    \caption{Correlation plot of Baichuan-7B. When the sequence length exceeds the pre-training length of 4k, the dominant dimension of Baichuan exhibits a sudden change.}
    \label{fig:baichuan_ood}
\end{figure}

\subsection{Correlation plot on Chinese datasets}
\label{corrrlation_chinese}
\begin{figure}[ht] 
    \includegraphics[width=\columnwidth]{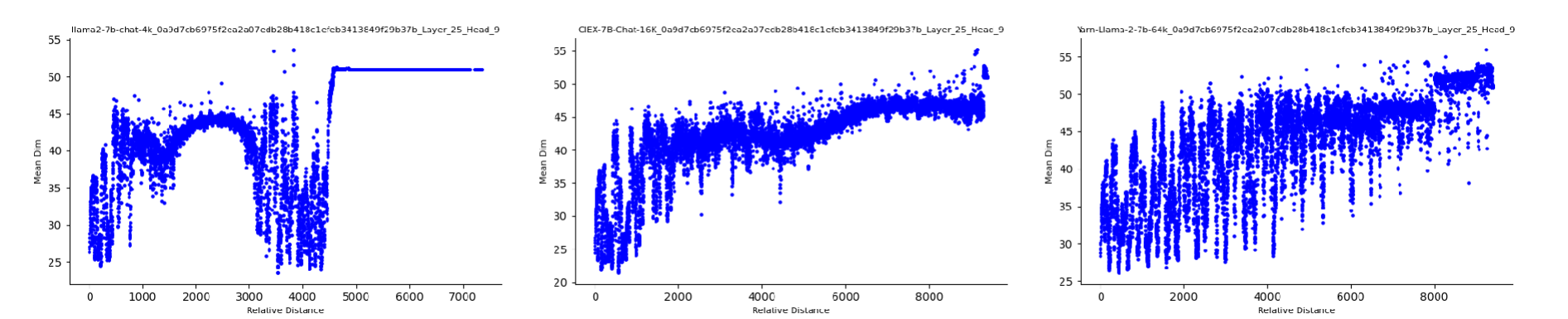} 
    \caption{Correlation plot comparing the original Llama model with two different length extrapolation methods on Chinese datasets.}
    \label{fig:correlation_chinese}
\end{figure}

\subsection{Spearman Correlation Coefficient in Different Methods}
\label{spearman_all}
\begin{figure}[H]
    \includegraphics[width=\columnwidth]{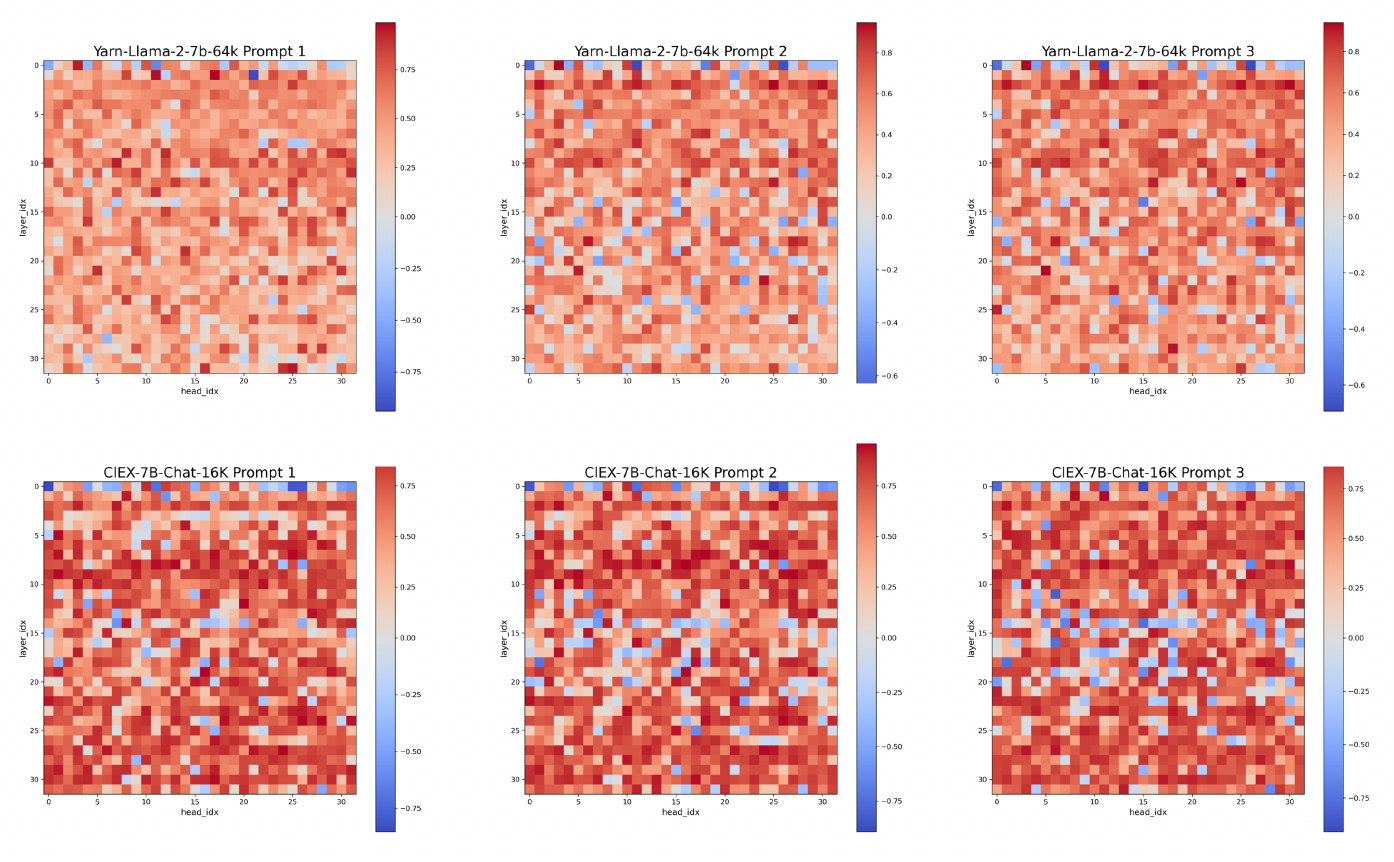}
    \caption{Spearman correlation coefficients of each head in the YaRN-Llama-2-7b-64K model and CLEX-7B-chat-16K across different prompts, illustrating the stability of position heads across prompts. In most heads, there is a notable correlation between the dominant dimension and the relative distance.}
    \label{fig:spearman_all}
\end{figure}

\subsection{Other masking Results}
\label{other_mask_result}
\begin{table*}[ht]
    \centering
    \small 
    \setlength{\tabcolsep}{3pt} 
    \renewcommand{\arraystretch}{1.1} 
    \begin{tabular}{|c|c|c|c|c|c|c|c|c|c|c|c|c|}
        \hline
        & \multicolumn{3}{c|}{Qasper} & \multicolumn{3}{c|}{MultifieldQA} & \multicolumn{3}{c|}{HotpotQA} & \multicolumn{3}{c|}{2WikiMQA} \\
        \hline
        masking method & 0-4k & 4-8k & 8k+ & 0-4k & 4-8k & 8k+ & 0-4k & 4-8k & 8k+ & 0-4k & 4-8k & 8k+ \\
        \hline
        Selfextend-no-masking & 19.52 & 16.27 & 21.39 & 40.73 & 34.77 & 27.25 & 45.5 & 41.86 & 40.21 & 40.12 & 32.64 & 28.07 \\
        \hline
        Selfextend-Random5\% & 14.05 & 16.07 & 5.33 & 37.83 & 24.57 & 23.47 & 42.8 & 39.73 & 36.97 & 39.49 & 33.14 & 22.11 \\
        \hline
        Selfextend-Random10\% & 15.19 & 15.08 & 3.95 & 37.57 & 23.81 & 18.97 & 44.03 & 39.42 & 31.71 & 33.94 & 29.63 & 20.45 \\
        \hline
        Selfextend-Top5\% & 17.43 & 12.27 & 4.59 & 37.74 & 23.52 & 20.62 & 42.53 & 16.17 & 7.91 & 29.19 & 17.85 & 6.46 \\
        \hline
        Selfextend-Top10\% & 8.19 & 7.39 & 3.75 & 31.92 & 17.89 & 14.28 & 33.39 & 14.57 & 5.54 & 30.7 & 12.12 & 5.15 \\
        \hline
        CLEX-no-masking & 25.06 & 27.69 & 19.94 & 48.31 & 32.88 & 24.75 & 21.42 & 23.88 & 28.0 & 21.76 & 20.55 & 9.01 \\
        \hline
        CLEX-Random5\% & 21.52 & 26.12 & 15.88 & 43.43 & 31.52 & 24.0 & 24.22 & 17.87 & 22.22 & 19.92 & 18.58 & 11.02 \\
        \hline
        CLEX-Random10\% & 22.55 & 27.15 & 19.7 & 46.94 & 30.56 & 19.75 & 25.85 & 24.65 & 29.59 & 19.0 & 18.5 & 14.29 \\
        \hline
        CLEX-Top5\% & 17.59 & 22.34 & 12.04 & 42.57 & 31.13 & 34.73 & 23.22 & 24.61 & 27.17 & 21.77 & 18.36 & 14.44 \\
        \hline
        CLEX-Top10\% & 13.64 & 18.82 & 8.66 & 45.34 & 29.56 & 22.63 & 21.49 & 22.18 & 24.26 & 17.99 & 20.05 & 12.83 \\
        \hline
    \end{tabular}
    \caption{Self-extend performance on QA tasks when masking out heads with top scores vs. random heads. Removing heads with top scores significantly reduces performance.}
    \label{tab:selfextend-mask-qa}
\end{table*}

\end{document}